\documentclass[letterpaper, 10 pt, conference]{ieeeconf}  

\IEEEoverridecommandlockouts                              

\usepackage{amsmath} 
\usepackage{amssymb}  
\usepackage{tikz}
\usetikzlibrary{positioning, arrows.meta, quotes}
\usetikzlibrary{shapes,snakes}
\usetikzlibrary{bayesnet}
\tikzset{>=latex}
\tikzstyle{plate caption} = [caption, node distance=0, inner sep=0pt,
below left=5pt and 0pt of #1.south]
\usepackage{algorithmic}
\usepackage{graphicx}
\usepackage{subcaption}
\usepackage{changepage}
\usepackage{textcomp}
\usepackage{xcolor}
\usepackage{url}
\usepackage{multirow}
\usepackage[bookmarks=true]{hyperref} 
\usepackage{array}
\usepackage{booktabs}
\usepackage{tabularray}
\usepackage{makecell}
\usepackage{times}
\usepackage{comment}
\usepackage{cite}

\title{Design and Evaluation of a Touchscreen-Based Teleoperation Interface \\ for Robotic Manipulators}

\author{Juan José García Cárdenas$^{1}$ $^{\star}$, Alperen Kenan$^{2}$ $^{\star}$, Hamidreza Raei$^{3}$ $^{\star}$, \\ Paul Bremner$^{2}$, Manuel Giuliani$^{4}$, Arash Ajoudani$^{3}$ and Adriana Tapus$^{1}$
\thanks{$^{\star}$These authors contributed equally to this work.}
\thanks{*This work was supported by the European Commission’s Marie Skłodowska-Curie Actions (MSCA) Project RAICAM (GA 101072634), and UK Research and Innovation (UKRI) grant number EP/X025977/1.}
\thanks{For the purpose of open access, the author has applied a Creative Commons Attribution (CC BY) license to any Author Accepted Manuscript version arising.}
\thanks{$^{1}$Juan José García Cárdenas is a PhD candidate and Adriana Tapus is Full Professor in the Computer Science and System Engineering Department (U2IS), Autonomous Systems and Robotics Lab, ENSTA, Institut Polytechnique de Paris, Paris, France {\tt\small juan-jose.garcia@ensta.fr; adriana.tapus@ensta.fr}$^{2}$Alperen Kenan is a PhD candidate and Paul Bremmer is an Associate Professor in Human-Robot Interaction at the Bristol Robotics Laboratory, University of the West of England, Bristol, United Kingdom {\tt\small alperen.kenan@uwe.ac.uk, paul2.bremner@uwe.ac.uk} $^{3}$Hamidreza Raei is a PhD candidate and Arash Ajoudani is a Full Professor in IIT, Genova, Italy {\tt\small hamidreza.raei@iit.it, arash.ajoudani@iit.it} $^{4}$Manuel Giuliani is a Professor at Kempten University of Applied Sciences, Kempten, Germany {\tt\small manuel.giuliani@hs-kempten.de}}}

\begin{document}

\maketitle
\thispagestyle{empty}
\pagestyle{empty}

\begin{abstract}

Intuitive teleoperation interfaces are crucial for the safe and effective operation of robotic manipulators in challenging environments. In the nuclear industry, surface contact tasks such as swab sampling require precise path and force tracking, obstacle avoidance, and sustained operator attention, which conventional joystick interfaces struggle to support effectively. 
This study designs and evaluates a novel touchscreen teleoperation interface that maps continuous finger movements directly to robotic manipulator motions, provides finer velocity control, and integrates control with visualization, enabling more natural, precise, and intuitive surface interaction than conventional controllers. A comparative user study with 20 participants evaluated task performance and workload using the proposed touchscreen, a conventional joystick, and a single-click autonomous mode. Tasks simulated realistic surface manipulation using a Franka Emika Panda arm, remotely controlled from another country. Kinematic, physiological, and behavioral data were recorded to comprehensively assess task performance, cognitive load, and operator trust across each control condition.
Participants completed teleoperation tasks more efficiently and accurately with the touchscreen interface, achieving a 53.5\% reduction in completion time (median: 2.50 vs. 5.38 min), higher in-area coverage on the sinusoidal path (90.7\% vs. 84.1\%), and lower overshoot on both path geometries compared with the joystick. Cognitive load, quantified via NASA-TLX (0–100), decreased from joystick to touchscreen (mean TLX 52 to 43; -9 points, -17.3\%) and was lowest under the autonomous one-click mode (31; -21 points vs. joystick, -40.4\%; -12 vs. touchscreen, -27.9\%). This research presents an easy-to-implement touchscreen interface that improves performance in teleoperated surface tasks while reducing cognitive load, providing a practical approach to safer and more effective manipulation in challenging environments.

\end{abstract}

\section{Introduction} 

Teleoperation of robotic manipulators is critical in hazardous and inaccessible environments, such as nuclear power plants (NPPs), and has long been endorsed by the International Atomic Energy Agency (IAEA) as a best practice strategy to reduce radiation exposure and human error during maintenance, inspection, and decontamination tasks \cite{iaea}. Surface interaction tasks, such as swabbing, play a critical role in maintenance within the nuclear industry \cite{johnson2022swab}. These tasks demand optimized contact force application, accurate path following, and sustained operator attention \cite{johnson2021force}. However, current teleoperation interfaces rely heavily on operator skill to achieve high performance, requiring substantial training and experience to maintain consistent results, particularly when precise force control is necessary \cite{kenan2025robot}.

In surface interaction, the operator must simultaneously regulate contact force normal to the surface while executing smooth tangential motion along the surface. This coupling makes contact tasks significantly more demanding than free-space teleoperation, because small velocity errors can lead to force oscillations, surface detachment, or excessive contact forces. An interface that reduces this burden must therefore support contact rich telemanipulation, not only intuitive motion input.

In surface interaction tasks, operators must control both normal contact force and smooth tangential motion. This coupling makes them more demanding than free-space teleoperation, as small velocity errors can cause force oscillations, detachment, or excessive forces. Effective interfaces must therefore support contact-rich telemanipulation, not just intuitive motion input.

Common joystick interfaces require operators to memorize button functions, operate the robot while monitoring a separate display, and mentally map the angular position of the joystick to the end-effector velocity \cite{kenan2025robot}. Joysticks also offer limited constant velocity control at their extremes, making it difficult to maintain continuous paths while maintaining a uniform force profile across a surface. The prevalence of touchscreen devices (phones/tablets) has introduced the potential for touchscreen-based teleoperation interfaces that provide intuitive, portable, and low-cost alternatives to conventional control devices. Touchscreens can provide a direct and flexible interaction modality by mapping human gestures to robotic motion, potentially reducing cognitive load and improving task performance. However, their adoption in teleoperation for precise surface interaction tasks remains limited, with most implementations relying on buttons, sliders, and on-screen menus, or for multi-touch gesture functions such as rotation or zoom \cite{cho2018assisted, paravati2011multi_touch}, rather than for direct, natural, one-to-one positional control of manipulators. Moreover, many touchscreen teleoperation approaches focus on free space positioning, waypoint/trajectory specification, or supervisory control, and do not explicitly target stable surface contact with controlled normal forces under real operational constraints (camera feedback, remote communication, and limited bandwidth).
Many touchscreen teleoperation approaches focus on free-space positioning, waypoint specification, or supervisory control, rather than maintaining stable surface contact with controlled normal forces under real operational constraints.

To address this gap, a novel touchscreen teleoperation interface is presented that maps finger movements directly to robotic manipulator motions. Different from prior touch-based interfaces that primarily address motion input, our system targets contact-rich surface manipulation by coupling collocated touch mapping with a control scheme that enforces a desired normal-force profile while maintaining compliant tangential tracking along the surface.
The interface aims to: (i) reduce operator workload through intuitive control mappings that emulate direct interaction with the environment; (ii) improve task performance and consistency by enabling continuous tangential path execution while autonomously tracking a target normal-force profile during surface manipulation; and (iii) enhance operator safety by enabling remote operation in hazardous environments. 

\begin{figure}[t]
\centering
\includegraphics[width=1.00 \columnwidth]{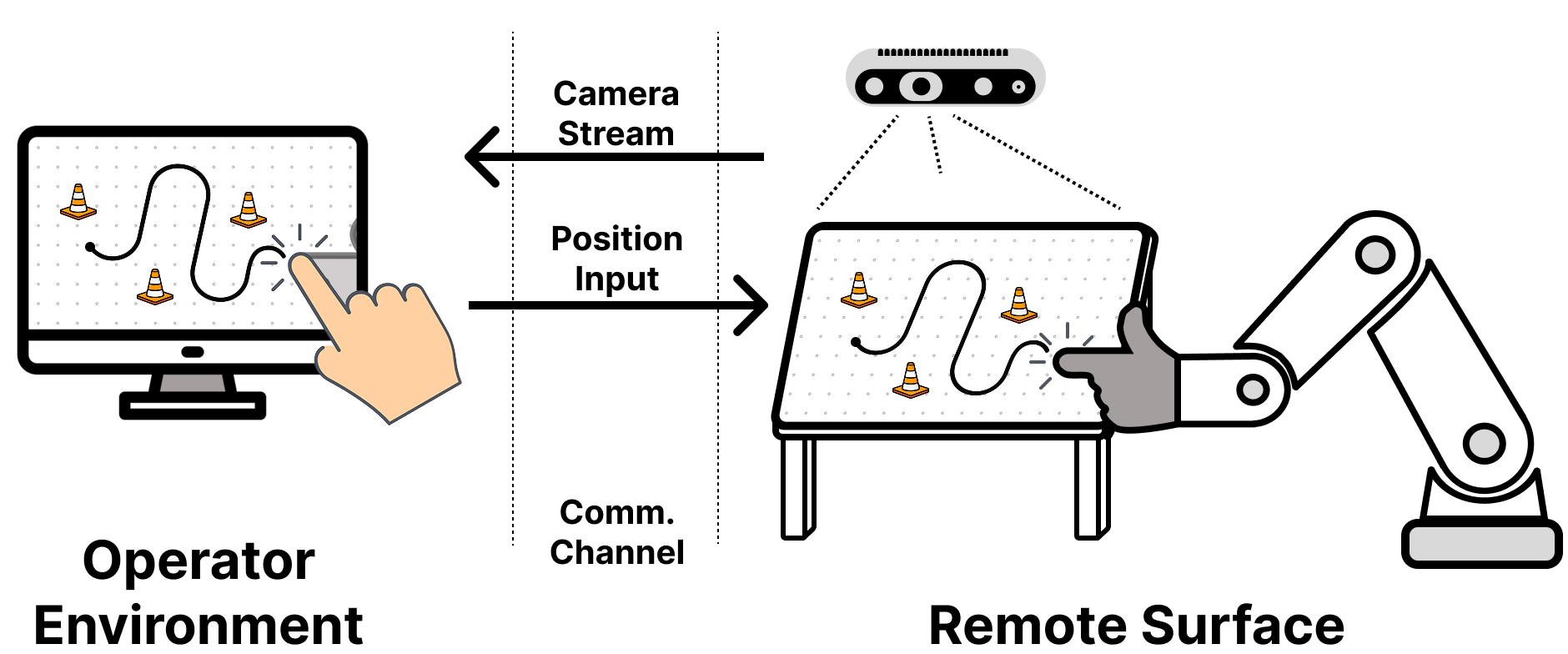}
\caption{Illustration of the touchscreen teleoperation interface}
\label{fig:illustration}
\vspace{-15pt}
\end{figure}

Figure \ref{fig:illustration} illustrates the working principle of the touchscreen teleoperation interface. The planar position of the operator's finger on the screen is mapped to the robot end-effector coordinates to interact with the surface, while the camera stream of the surface is displayed on the screen to enhance the intuitiveness of the robot control, giving the operator the impression of direct contact with the environment. Critically, while the operator specifies tangential motion through touch, the robot autonomously regulates the normal contact force, reducing the need for continuous manual force modulation during swabbing tasks.

This paper makes three primary contributions: (1) the design of a novel touchscreen interface for teleoperation of robotic manipulators in surface interaction tasks; (2) the implementation of a hybrid impedance force controller that tracks a force profile in the normal direction while allowing compliant position tracking in tangential directions; and (3) user evaluations with 20 participants, providing experimental results on the potential of touchscreen-based teleoperation.

The remainder of the paper is organized as follows. Section \ref{sec:relatedwork} reviews prior work on teleoperation interfaces used in hazardous environments and on physiological estimation of cognitive load and trust. Section \ref{sec:Methodology} describes the experimental methodology, including hardware architecture, network setup, and sensor pipeline. Section \ref{sec:teleoperation} details the control algorithm for robotic teleoperation and manipulator-surface interaction. Section \ref{sec:Hypothesis} outlines three hypotheses on the relationships among autonomy level, cognitive load, and trust. Section \ref{sec:Results} reports quantitative results on task performance, operator workload, and trust dynamics. Section \ref{sec:Discussion} discusses cross-border teleoperation deployments, study limitations, and future work. Finally, Section \ref{sec:Conclusions} summarizes the main findings and their implications for touchscreen-based teleoperation in precision surface-interaction tasks.

\section{Related Work}
\label{sec:relatedwork}

Teleoperation of robotic systems in challenging or restricted environments, such as nuclear, aerospace, medical, and industrial settings, has played a key role in reducing human exposure to risk while ensuring task precision. Early deployments of robotic manipulators in nuclear facilities demonstrated that remotely operated systems can reduce radiation exposure and costs by performing inspection and maintenance tasks in hazardous environments \cite{bogue2011robots}. Later reviews of inspection systems \cite{sundar2012design} and robotic strategies in nuclear environments \cite{smith2020robotic} highlighted key issues such as deployment constraints, while stressing the need for robust and adaptable platforms. These examples highlight the continuing importance of advanced robotic teleoperation for the next generation of nuclear maintenance. In high-risk settings, Grobbel et al. \cite{Grobbel2025} proposed a universal shared-control framework that adapts in real time to operator intentions without explicit force cues . Cheng et al. \cite{Cheng2024} introduced an efficient shared-control method for grasping that adapts to unpredictable and underspecified telemanipulation tasks, improving operational flexibility. These studies highlight the need for robust, scalable, and latency-resilient shared-control systems that preserve human oversight while reducing the burden of continuous manual input.

Interface design plays a crucial role in enabling precise and cognitively efficient teleoperation of robotic platforms. Recent studies have explored spatial and multimodal interfaces to improve operator situational awareness. Smith and Kennedy \cite{Smith2024} introduced an augmented reality (AR) interface that allows users to directly manipulate robotic end-effectors in 3D space with intuitive depth cues, reducing cognitive load and improving manipulation accuracy . Lii et al.~\cite{Lii2022} presented the Exodex Adam, a reconfigurable haptic device enabling full-hand kinesthetic interaction for immersive robot control. Su et al.~\cite{Su2023} surveyed the use of Extended Reality (XR) technologies, including virtual, augmented, and mixed reality, for telemanipulation in hazardous industrial contexts, showing how these interfaces reduce user strain and improve task performance. These advances reflect a trend toward more embodied and adaptive interfaces that extend operator capabilities beyond conventional screens and joysticks.

Touch-based interfaces for robot control have been explored for over two decades in supervisory and direct-manipulation settings. Hwang et al. \cite{hwang2003fingertip} used fingertip input to generate smooth Bézier trajectories for supervisory control of mobile robots, demonstrating the value of sketch-like interaction for motion specification . For manipulators, Farkhatdinov et al. compared indirect vs. direct workspace representations on a touchscreen and showed that displaying the task space directly on the screen can improve usability for end-effector control \cite{farkhatdinov2009workspace}. 
Toh et al. \cite{toh2012multitouch} proposed a multi-touch interface for dexterous telemanipulation, showing that multi-touch gestures can encode richer kinematic commands than traditional single-axis devices . Singh et al. \cite{singh2013fatigue} presented a remote manipulator interface aimed at reducing task load and fatigue, reinforcing that touch-based interaction can reduce operator burden in teleoperation tasks. Touchscreens have also been used to capture human motion for offline robot learning, with finger trajectories recorded to teach robots human-like handwriting motions \cite{kenan2026handwriting}. In contrast to much of this prior work, often focused on free space motion, trajectory specification, kinematic command richness, or offline motion learning, our study targets contact surface interaction where stable execution requires simultaneous tangential motion and explicit regulation of the normal contact force. We, therefore, combine a collocated touch mapping with a controller that tracks a desired normal-force profile while maintaining compliant tangential tracking, and we evaluate this design under realistic remote-operation constraints with performance metrics and multimodal workload/trust measures.
We combine collocated touch mapping with a controller that maintains desired normal force and compliant tangential tracking, and evaluate it under realistic remote-operation constraints using performance and multimodal workload/trust metrics.

Finally, trust and cognitive load are critical factors in Human-Robot Interaction (HRI), influencing both performance and user willingness to rely on automation. Physiological sensing has emerged as a powerful method for estimating these variables in real time. Shared-control systems have been shown to reduce workload and moderately enhance trust, as measured by pupil diameter and the NASA-TLX \cite{Pan2024}. Wang et al.~\cite{Wang2024} profiled cognitive load at high resolution using multimodal signals such as electrodermal activity (EDA), heart rate variability, and blink dynamics. In industrial HRI, Campagna et al.~\cite{Campagna2024} used GSR alone to detect trust levels during safety-critical handovers with 69\% accuracy, while Yang et al.~\cite{Yang2024} developed an adaptive semi-autonomous surgical tool guided by real-time workload estimates. Zakeri et al.~\cite{Zakeri2023} combined EEG, behavioral, and physiological signals for workload assessment in smart factory teleoperation, showing that cross-modal sensing improves estimation fidelity. This study builds on previous work by fusing blink rate, GSR, and thermal imaging with behavioural and performance metrics to track trust and cognitive load in robotic teleoperation. Unlike prior approaches, the proposed framework integrates real-time physiological sensing with a touchscreen-based interface, shared autonomy, and task performance metrics, providing a unified experimental platform for adaptive HRI. 

While previous studies have investigated alternative teleoperation-based interfaces and shared-control approaches, they often fall short of providing intuitive, effective, low-effort operation in hazardous environments. This work presents a novel touchscreen-based interface that directly maps finger movements to robotic manipulator motions during surface interactions, supported by a hybrid impedance force controller. The system also incorporates multimodal physiological sensing to assess operator cognitive load across different interface conditions. Together, this unified framework offers an accessible alternative to traditional joystick control while promoting safe, effective, and cognitively efficient teleoperation.

\section{Methodology}
\label{sec:Methodology}

\begin{figure}[ht]
\centering
\includegraphics[width=1.00 \columnwidth]{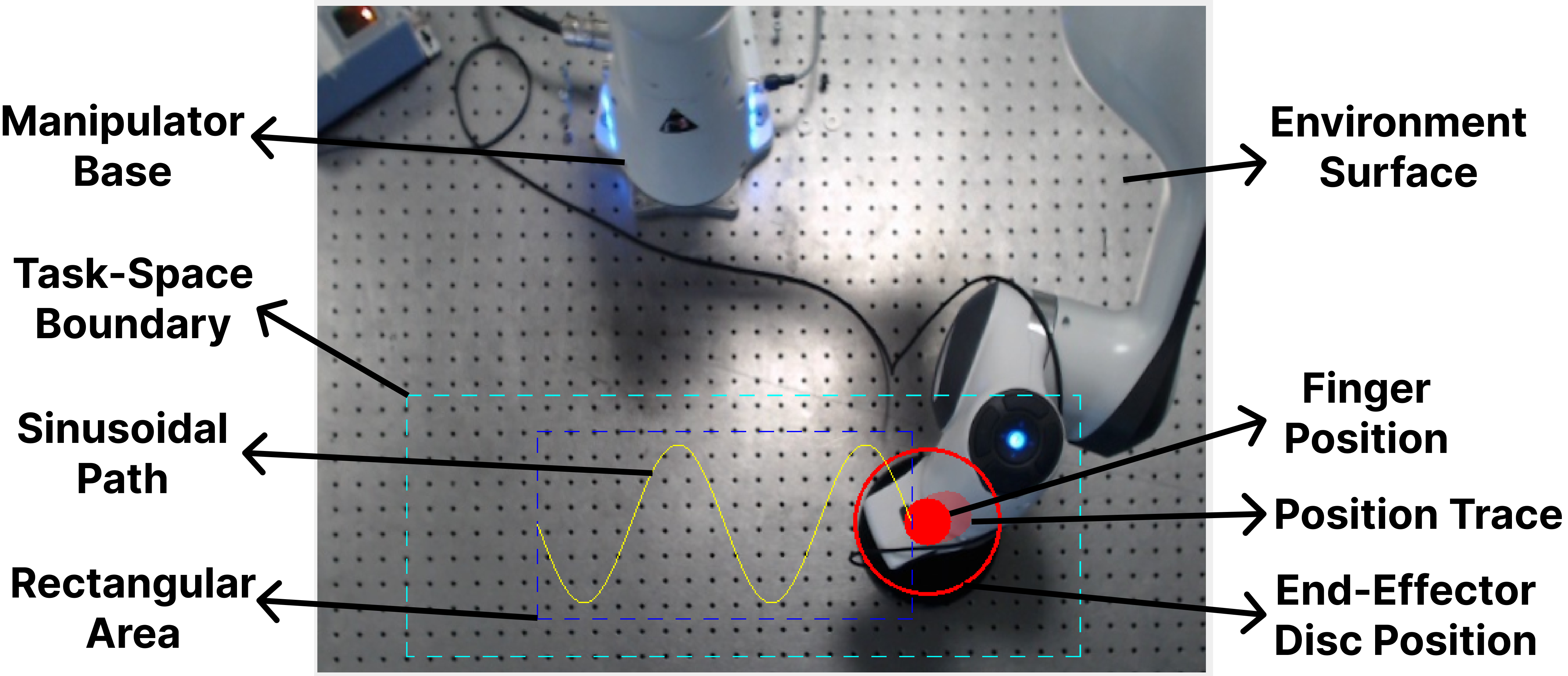}
\caption{Interface display captured during trial execution}
\label{fig:ss}
\vspace{-15pt}
\end{figure}

To evaluate the proposed touchscreen-based interface for robot teleoperation, a comparative user study was conducted against an industry-standard joystick controller. The objective was to assess usability, cognitive load, and task performance. Tasks were modelled on realistic scenarios such as swabbing in nuclear maintenance, designed to replicate precise, contact-based manipulation over a surface. Figure \ref{fig:ss} displays a full-screen screenshot of the actual interface used by the participants. The interface provides a real-time top-view video feed of the manipulator. The red dot indicates the operator’s finger position, which is sent to the manipulator as a position reference, while the surrounding red circle shows the manipulator’s end-effector position. Presenting the finger input and manipulator video feed output on the same screen allows operators to control the robot directly from the display they are observing. The blue dashed line indicates the manipulator’s virtual boundaries, and the yellow dashed line represents the sinusoidal path to be followed. Previous positions leave a trace, and the bottom-left corner of the screen displays the video feed’s frame rate.

\subsection{Participants}

A total of 20 participants took part in the study, of whom 17 chose to disclose demographic information. Among these, 11 were identified as male, 5 as female, and 1 opted not to reveal their gender. Participant ages ranged from 19 to 67 years (\emph{M} = 32.12, \emph{SD} = 11.17). The group was geographically diverse, with 10 participants from Europe, 4 from Asia, 1 from Africa, and 2 from South America. Educational backgrounds included 5 PhD students, 8 Master students, and 4 undergraduates. Most participants reported some prior experience with robots (19 out of 20) and considered themselves proficient with controllers or simulators (18 out of 20). All expressed a positive outlook on the growing role of robots in daily life and workplace tasks.

\subsection{Study Environment and Context}

The study took place at the Bristol Robotics Laboratory (Bristol, UK), where participants interacted with the interface, while the teleoperated robot manipulator was located remotely at IIT (Genova, Italy) (see Figure \ref{fig:vpn}). 

\begin{figure}[ht]
    \centering
    \begin{subfigure}[b]{0.48\columnwidth}
        \centering
        \includegraphics[width=\linewidth]{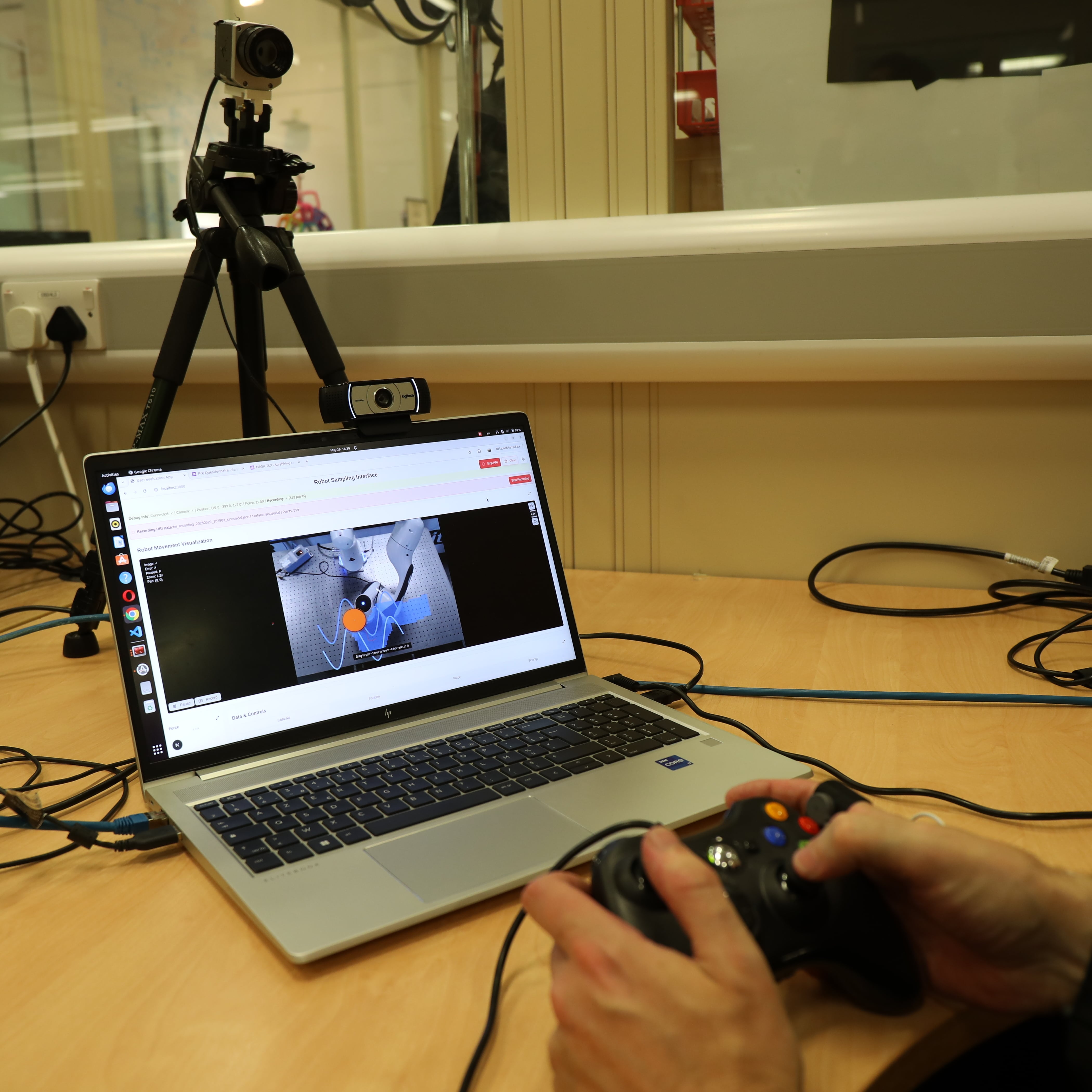}
        \caption{Joystick controller}
        \label{fig:subfig1}
    \end{subfigure}
    \hfill
    \begin{subfigure}[b]{0.48\columnwidth}
        \centering
        \includegraphics[width=\linewidth]{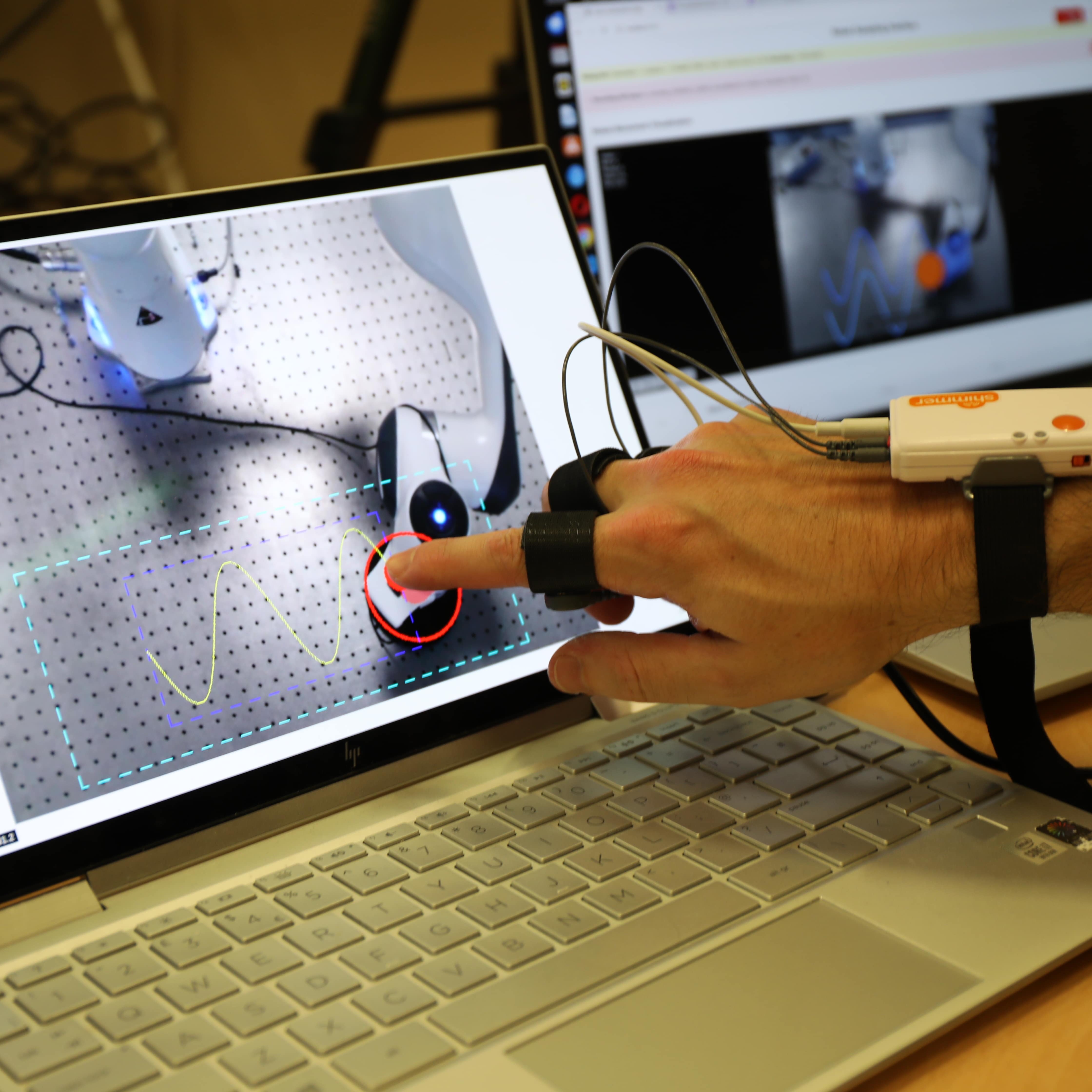}
        \caption{Touchscreen controller}
        \label{fig:subfig2}
    \end{subfigure}
    \caption{Two manual control conditions tested by participants.}
    \label{fig:test_conditions}
    \vspace{-10pt}
\end{figure}

This remote setup simulated realistic teleoperation scenarios in which the communication channel may be less stable than direct cable connections, and operators may control robots over long distances in safety-critical environments. Before the sessions, participants completed a questionnaire on age, gender, academic background, prior robotics exposure, and weekly gaming/joystick hours, following \cite{Cheng2024}. Each session started with a study introduction, followed by a brief tutorial and a familiarization period, allowing participants to gain confidence before proceeding.
Teleoperation tasks were performed under three conditions: (C1) using a conventional Xbox joystick, (C2) using the custom touchscreen-based interface, and (C3) a supervised single-click autonomous playback mode, included as a controlled lower bound on cognitive load rather than a deployable alternative, playback presupposes a fully known, static task, whereas realistic decontamination requires online operator decisions. All control conditions used identical visual feedback to isolate camera feed effects, and the two manual control conditions are shown in Figure \ref{fig:test_conditions}. 

The joystick controller and keyboard were standard commercial products. The touchscreen-based interface was implemented on a 13-inch HP Spectre x360 Convertible laptop with an ELAN2514 touch controller. The experiments employed a Franka Emika Panda robotic arm fitted with an ATI Mini45 6-axis force/torque sensor to perform swabbing over a flat surface. An Azure VPN connected the operator station in Bristol, UK to the robot site in Genoa, Italy, streaming ROS topics with a verified round-trip latency below 100 ms. 

Twenty participants each executed the prescribed swabbing pattern in all three control modes. During each run, the system recorded manipulator kinematics, end-effector position/force, joystick/touchscreen inputs, task completion time, and live overhead video with the target trajectory projected for performance evaluation. Besides, physiological signals and cognitive activity were recorded for each user, which will be explained in more detail in the subsection below. In all conditions, participants were instructed to follow a sinusoidal path and to fill a rectangular area, simulating a surface coverage task. The order of conditions and task paths was 
randomized across participants to minimize learning effects. 
To reduce potential bias, participants were not informed of the study’s hypotheses. The open-source code for this project is publicly available.\footnote{The repository can be accessed at: \url{https://github.com/kenanalperen/Touch_Screen_Interface.git}.
It includes ROS-based Python nodes for finger tracking, joystick control, angle calibration, GUI visualization, trajectory generation and documentation of the setup.}

\subsection{Cognitive Load and Trust Evaluation}

To assess cognitive workload and changes in operator trust, both qualitative and quantitative data were collected. After each experimental condition, participants completed the NASA Task Load Index (TLX) \cite{nasatlx}, providing six subjective workload sub-scores. Following their final run, participants also offered open-ended feedback on interface usability, perceived autonomy, and trust. These were complemented by physiological measures including blink rate, facial temperature, and GSR, which can reflect cognitive workload and stress \cite{garcia2025legged} \cite{garcia2024}. Together, these data provided a multimodal record of operator workload alongside performance metrics.
High-resolution physiological and behavioral signals were captured by combining RGB video, thermal imaging, and electrodermal sensing. Facial features were recorded with a Logitech C930e RGB webcam (1920 × 1080 px, 30 fps) \cite{logitechC930e} along with the OpenFace ROS node for AU45 extraction, and an Optris PI-640 thermal camera (640 × 480 px, 32 fps, -20 °C - 100 °C) \cite{optrisPI640}. Thermal frames were time-stamped and later mapped to the 68 dlib facial landmarks to extract mean temperature in the nose, periorbital, and forehead regions \cite{dlib}. Skin-conductance activity was monitored with Shimmer 3 GSR+ \cite{shimmerGSR3} units affixed to the index and middle fingers of the dominant hand. The sensors streamed conductance at 128 Hz, a rate recommended for detecting fast, phasic responses that track moment-to-moment cognitive demands \cite{boucsein2012electrodermal,critchley2002electrodermal}. GSR provides a non-invasive proxy for sympathetic arousal \cite{berntson2007cardiovascular}. All feeds fed a custom dashboard (live video, thermal ROIs, GSR traces, blink counts) for online monitoring and on-site annotation of high-load events. This synchronized pipeline yielded, per condition, a Cognitive-Load Index (CLI) and a Trust Index (TI). The CLI fuses phasic GSR peak rate, blink rate, and nose/periorbital/forehead temperature deviations after per-participant z-normalization. The TI is not a single-channel measure: trust is estimated with the validated multimodal Bayesian fusion model from our recent study \cite{garciaCardenas2025DynamicTrust} (building on Guo and Yang \cite{Guo2021BayesianTrust}), in which task-performance and z-normalized physiological features jointly drive the posterior reported in Section \ref{sec:Results}

\subsection{Ethics}

This study was approved by the Research Ethics Committee of the University of the West of England, College of Arts, Technology \& Environment (Reference: 13470965). 
A participant information sheet was provided to all participants prior to their consent to take part, and signed consent forms were collected before the experiments started. Appropriate measures were taken to ensure participant confidentiality and data security throughout the study. Participants had the right to withdraw at any point and to request the removal of their data up to seven days following the study.

\section{Teleoperation Control Implementation}
\label{sec:teleoperation}

\begin{figure}[ht]
\centering
\includegraphics[width=1.00 \columnwidth]{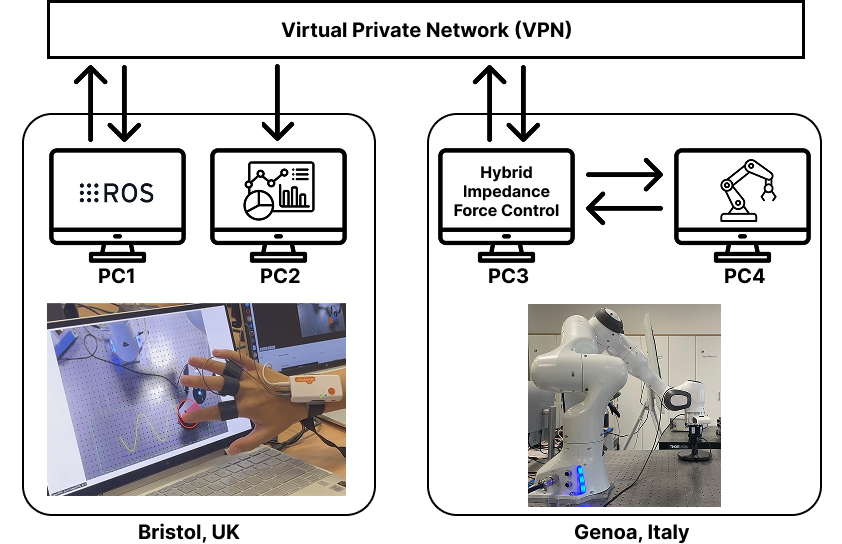}
\caption{Overview of the communication framework}
\label{fig:vpn}
\vspace{-10pt}
\end{figure}

The communication architecture for the teleoperation setup is illustrated in Figure \ref{fig:vpn}, which involves four PCs located in different countries. In Bristol, UK, PC1 served as the human-robot interface, capturing finger position data and displaying the robot’s camera feed, while PC2 logged experimental data. In Genoa, Italy, PC3 converted position reference data into a time-synchronised ROS topic, and PC4 acted as the ROS Master, controlling the Franka Emika Panda arm and enforcing safety constraints. All PCs were connected to the same virtual network via a VPN, with wired links ensuring reliable performance.

\subsection{Touchscreen Input Conversion}

To convert pixel coordinates from the touchscreen into physical world coordinates, the pixels-per-millimeter (PPM) metric is first calculated using the screen width \(W_{\mathrm{px}}\), screen height \(H_{\mathrm{px}}\) in pixels, and the screen diagonal length \(D_{\mathrm{mm}}\) in millimeters, as follows:
 
\begin{align}
    \mathrm{PPM} = \frac{\sqrt{W_{\mathrm{px}}^{2} + H_{\mathrm{px}}^{2}}}{D_{\mathrm{mm}}}
\end{align}

Given a screen coordinate \((u,v)\) in pixels, the corresponding raw physical coordinates \((x_{\mathrm{raw}}, y_{\mathrm{raw}})\) in millimeters are obtained by applying offset terms \(c_1\) and \(c_2\), and scaling factors \(k_1\) and \(k_2\):

\begin{align}
    x_{\mathrm{raw}} &= \left(\frac{v}{\mathrm{PPM}} + c_1 \right) \times k_1 \\
    y_{\mathrm{raw}} &= \left(\frac{u}{\mathrm{PPM}} + c_2 \right) \times k_2
\end{align}

In this implementation, scaling factors are identical in \(x\) and \(y\) directions, i.~e., \(k_1 = k_2\). The input coordinates are constrained within a predefined operational workspace. The robot’s current end-effector position \((x_{\mathrm{EE}}, y_{\mathrm{EE}})\) is obtained in millimetres, including offsets to define the initial reference point. The Euclidean distance between the clamped input position and the end-effector position is computed. If this distance is below a specified threshold \(d_{\max}\), the input position is published; otherwise, the end-effector position is retained to prevent sudden jumps in the input sent to the robot. Position and velocity constraints are enforced on the manipulator to ensure operational safety.

\subsection{Manipulator Control}

The control design of the manipulator addresses three constraints: (i) non-negligible end-to-end latency in the teleoperation link, (ii) strict regulation of the normal contact force to guarantee repeatable, verifiable swabbing, and (iii) mapping 2D touchscreen input to the motion of a 6-DoF end effector over planar, inclined, or curved surfaces. Therefore, a hybrid impedance-admittance architecture is adopted: the tangential directions ($x$–$y$) are governed by Cartesian impedance for compliant position tracking, while the normal direction ($z$) is regulated by an admittance loop that converts force error into a motion reference in the task/contact frame. A distal 6-DoF force/torque sensor closes the z-axis force loop locally at the robot (PC4), so stability-critical normal-force regulation is decoupled from network latency; only the tangential position reference traverses the link.
For contact-rich tasks, autonomy is integrated via a surface-lock behaviour: when the measured normal force in the end-effector frame exceeds a small threshold \textit{$F_{th}$}, the controller switches from free 3-D mapping to contact mode that sticks the tool to the surface. In this mode, the measured moments about the task-frame x-y axes (\textit{$M_x, M_y$}) are used to drive them toward zero, rotating the end effector until the task-frame \textit{$\hat{z}$} is aligned with the surface normal inferred from the wrench. The task-frame transformation is updated for swabbing at this point. A selection matrix defined in the end-effector/task frame enforces tangential/normal decoupling and prevents control overlap. The total joint torque is decomposed as shown in Equation \ref{eq:torque_overall}.

\begin{equation}
    \tau \;=\; \tau_{\text{xy}}^{\text{imp}} \;+\; \tau_{z}^{\text{adm}} \;+\; \tau_{\text{null}} \;+\; g(q),
    \label{eq:torque_overall}
\end{equation}

where $\tau_{\text{xy}}^{\text{imp}}$ enforces planar tracking, $\tau_{z}^{\text{adm}}$ enforces normal force regulation, $\tau_{\text{null}}$ preserves a comfortable posture without disturbing the task, and $g(q)$ is gravity compensation.

\subsubsection{Planar (\textit{x-y}) impedance}

Following Giammarino et~al.~\cite{giammarino2024super}, their Cartesian impedance formulation is adapted to the planar touchscreen teleoperation: the touchscreen provides $x$–$y$ position references in the task frame, and the resulting impedance term $\tau_{\text{xy}}^{\text{imp}}$ is computed accordingly.

\subsubsection{Normal (\textit{z}) admittance}
Let $q\in\mathbb{R}^n$ be the joint vector, $J_t(q)\in\mathbb{R}^{3\times n}$ the translational Jacobian expressed in the task/contact frame, and $S_f=\mathrm{diag}(0,0,1)$ the normal-axis selector. The desired normal trajectory $z_d$ is generated by a mass–spring–damper admittance driven by the force error (see Equation \ref{eq:pure_force})
\begin{equation}
    M_a\,\ddot z_d \;+\; B_a\,\dot z_d \;+\; K_a\,(z_d - z_0) \;=\; F_z^{*} - F_z.
    \label{eq:pure_force}
\end{equation}
and is tracked by a stiff inner loop producing a virtual normal wrench (see Equation \ref{eq:force_to_pos})
\begin{equation}
     F_z^{\text{pos}} \;=\; K_p^{z}\,(z_d - z) \;+\; K_d^{z}\,(\dot z_d - \dot z).
     \label{eq:force_to_pos}
\end{equation}
which is embedded along the task-frame $z$ axis and mapped to joints as shown in Equation \ref{eq:pos_to_torque}.
\begin{equation}
    \tau_{z}^{\text{adm}} \;=\; J_t^{\top}\, S_f \begin{bmatrix} 0 \\[1pt] 0 \\[1pt] F_z^{\text{pos}} \end{bmatrix}.
    \label{eq:pos_to_torque}
\end{equation}

\subsubsection{Adaptive force control}

At initial touch the controller executes a controlled force ramp from $1$\,N to $10$\,N over $1$\,s while sampling the normal force $F_z(t)$ and end–effector position $z(t)$ at $100$\,Hz. The normal indentation is defined relative to first contact as
$\delta(t) \!=\! z(t_0) - z(t)$ (so $\delta(t_0)\!=\!0$).
To account for mild nonlinearities in surface behavior, the force–displacement data are partitioned by force into three ranges, $\mathcal{S}_1\!: [1,3]$\,N, $\mathcal{S}_2\!: (3,7]$\,N, and $\mathcal{S}_3\!: (7,10]$\,N.
Within each range $\mathcal{S}_j$, a RANSAC regression is applied using the local model $F_i \approx k_j\,\delta_i$, producing the inlier set $\mathcal{I}_j \subset \mathcal{S}_j$. The resulting slope estimate of the local stiffness is shown in Equation \ref{eq:local_stiffness}.

\begin{equation}
    k_j \;=\; \frac{\sum_{i \in \mathcal{I}_j} F_i\,\delta_i}{\sum_{i \in \mathcal{I}_j} \delta_i^{\,2}} ,
    \label{eq:local_stiffness}
\end{equation}

The three local estimates are then aggregated into a single stiffness for control near the operating point (target force \(F_z^\star=5\,\mathrm{N}\)) via a convex weighted average, \(k_{\text{est}}=\sum_{j=1}^{3} w_j k_j\), with \(\mathbf{w}=[0.25,\,0.5,\,0.25]^{\top}\) and \(\sum_j w_j=1\).

The estimated stiffness of the surface is then utilized to tune the controller parameters as follows (see Equation \ref{eq:inner_gain_adaptation}):

\begin{equation}
    K_p^{z} \;=\; \alpha\,k_{\text{est}}, 
    \qquad
    K_d^{z} \;=\; 2\,\zeta\,\sqrt{M_a\,K_p^{z}},
    \label{eq:inner_gain_adaptation}
\end{equation}

with $\alpha\!>\!0$ a small scaling factor, $\zeta\!\in\![0.7,1]$ the desired damping ratio, and $M_a$ the (fixed) virtual mass from the admittance loop. This choice yields tight force regulation around $F_z^\star$ while maintaining well-damped transients across soft and hard surfaces. Practical saturations on $(z_d,\dot z_d)$ are retained to prevent excessive indentation without clipping corrective motion.

\section{Hypotheses}
\label{sec:Hypothesis}

Building on research in collaborative HRI, three different hypotheses are formulated: 

\begin{itemize}

    \item [\textbf{H1}] Participants in condition C3 (Supervisory one-click autonomy) will have the lowest cognitive load in comparison with the other two conditions of teleoperation. 
    \item [\textbf{H2}] Participants in condition C2 (Touchscreen interface) will elicit significantly lower workload and higher trust ratings than the baseline condition C1 (Joystick interface). 
    \item [\textbf{H3}] Participants, who are spatially skilled or have gaming experience, will display lower blink-rate elevations, smaller GSR spikes and shorter completion times than low-skill users in condition C1, with diminishing differences in conditions C2 and C3.
    
\end{itemize}

These hypotheses are motivated by prior research in HRI and human factors. For \textbf{H1}, delegating motion-planning, collision-avoidance, and force-control subtasks to a reliable autonomous controller reduces operator perceptual–motor and working-memory demands, resulting in lower NASA-TLX scores, reduced blink suppression, and decreased phasic-GSR peaks \cite{Sheridan1992SupervisoryControl,pruks2018shared_32,Guo2021BayesianTrust,Xu2020OPTIMO}. Prior shared- and full-autonomy studies consistently report that mental workload decreases monotonically with increasing autonomy as long as the system remains transparent and predictable \cite{hancock2011meta_36}. \textbf{H2} is supported by studies showing that touchscreen input shortens command chains, reduces manual effort, and enhances situational awareness compared with rate-controlled joysticks, while also providing richer contextual cues that help operators anticipate robot intent, fostering calibrated trust \cite{Green2008MultiTouchTeleop,rupp2013comparing_38,wu2024comparative_39,ding2022development_40,nenna2022human_41}. \textbf{H3} is based on evidence that spatial-orientation skill and gaming experience improve efficiency in perspective-taking tasks critical for indirect-view teleoperation, leading to fewer trajectory corrections, lower NASA-TLX ratings, and faster task completion, with differences diminishing as autonomy or interface support increases \cite{Hegarty2002SpatialOrient,nenna2022influence,Beatty1982PupilLoad}.

\section{Results}

\label{sec:Results}

\subsection{General performance}

\begin{figure}
    \centering
    \includegraphics[width=0.65\linewidth]{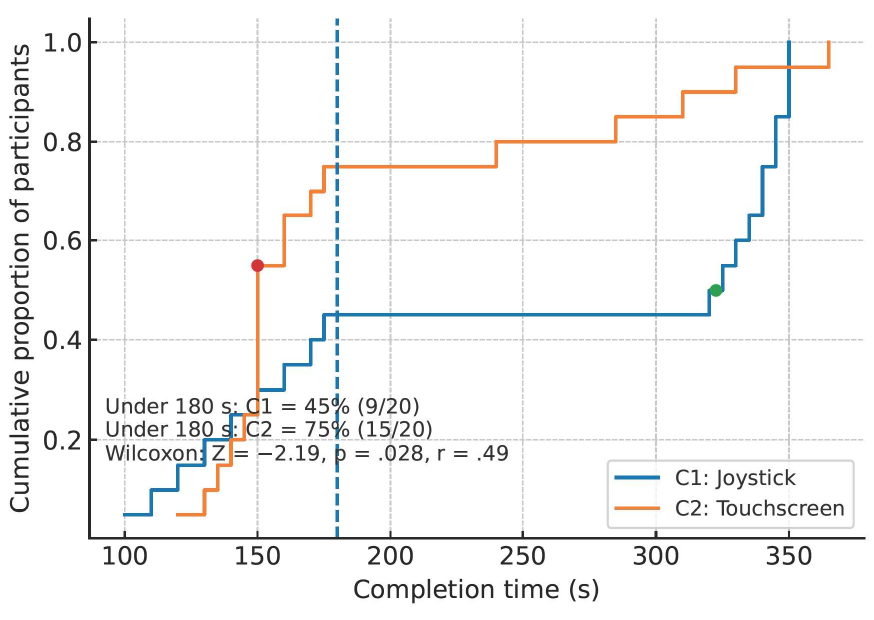}
    \caption{Completion-time ECDF by interface. The touchscreen (C2) shifts the distribution left relative to the joystick (C1); the dashed line marks 180\,s. Proportion under 180\,s: C1 = 45\% (9/20), C2 = 75\% (15/20).}
    \label{fig:ecdf-time}
    \vspace{-15pt}
\end{figure}

All 20 operators finished both manual conditions C1 and C2.
In terms of completion time, the touchscreen condition (C2) had a median of 2.50 min compared to 5.38 min with the joystick condition (C1), representing a 53.5\% reduction. Consistent with this, the ECDF (Empirical Cumulative Distribution Function) plot in Figure ~\ref{fig:ecdf-time} shows that the curve for C2 rises more steeply, corresponding to shorter completion times, whereas the curve for C1 is flatter and shifted to the right, indicating slower completion times. By 180s, 75\% of participants (15 out of 20) had completed the task with C2, compared to 45\% (9 out of 20) with C1. Completion times were significantly lower with C2, as confirmed by a Wilcoxon signed-rank test, which was chosen due to the non-normal distribution of the data ($Z=-2.19$, $p=0.028$, $r=0.49$). The autonomous one-click mode (C3) required no teleoperation and was excluded from this analysis.

The sampling disk ($\oslash$10 cm) traversed a target patch of $\approx 60 cm^2$ in two geometries: rectangular area and sinusoidal path. For the rectangular path, both interfaces achieved almost perfect in-area coverage (C1: \emph{M} = 98.2\%, \emph{SD} = 1.6; C2: \emph{M} = 99.3\%, \emph{SD} = 0.9; Wilcoxon, \emph{Z} = -1.54, \emph{p} = 0.12). In the more demanding sinusoidal path, average coverage improved from 84.1\% (\emph{SD} = 4.8) with C1 to 90.7\% (\emph{SD} = 3.9) with C2, an absolute improvement of 6.6\% (\emph{Z} = -2.87, \emph{p} = 0.004, \emph{r} = 0.64). In addition, the area outside the prescribed path was greatly reduced with C2 compared to C1. For the rectangle, mean excess surface declined from 17.3\% ($\approx 10.4 cm^2$) to 4.2\% ($\approx 2.5 cm^2$), and for the sinusoid from 14.1\% ($\approx 8.5 cm^2$) to 8.9\% ($\approx 5.3 cm^2$). Both reductions were statistically significant (rectangle: \emph{Z} = -3.63, \emph{p} $<$ 0.001; sinusoid: \emph{Z} = -2.15, \emph{p} = 0.031). Taken together, the results show that the touchscreen interface (C2) enabled faster task execution, higher in-area coverage, and substantially lower overshoot compared to the traditional joystick (C1).
Overall, the results indicate that the touchscreen-based interface (C2) facilitated faster task completion, greater in-area coverage, and markedly reduced overshoot compared to the traditional joystick (C1).

\subsection{Cognitive load and trust}

\begin{figure}[t]
  \centering
  \includegraphics[width=0.5\linewidth]{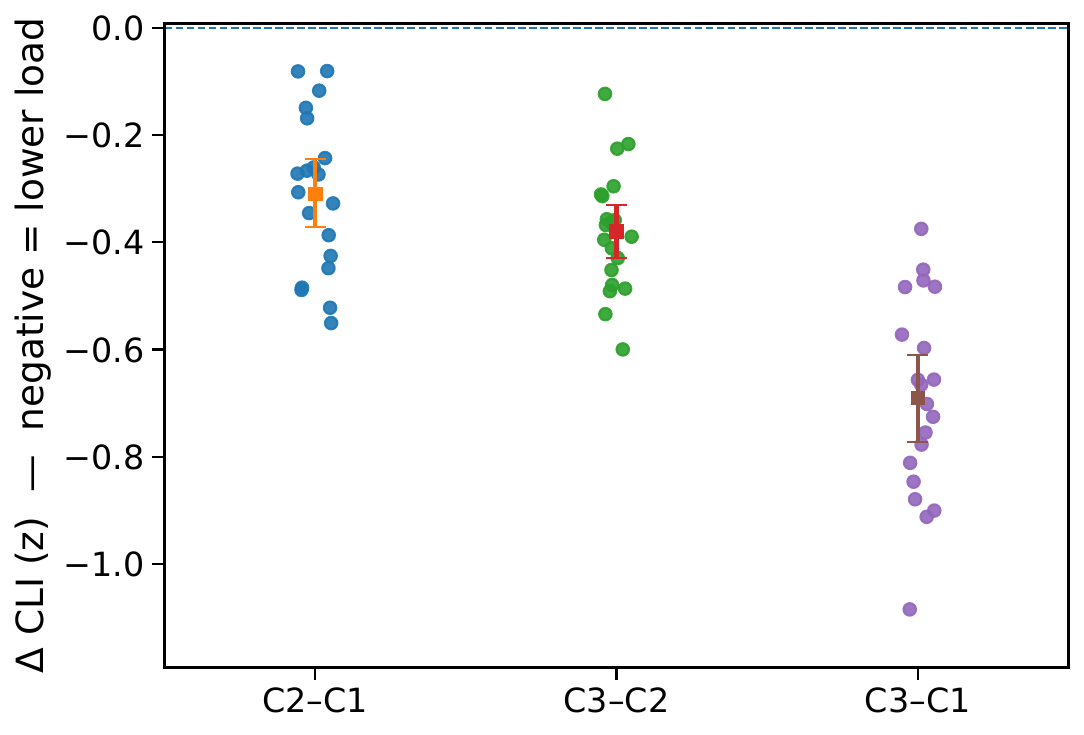}\\[4pt]
  \includegraphics[width=0.5\linewidth]{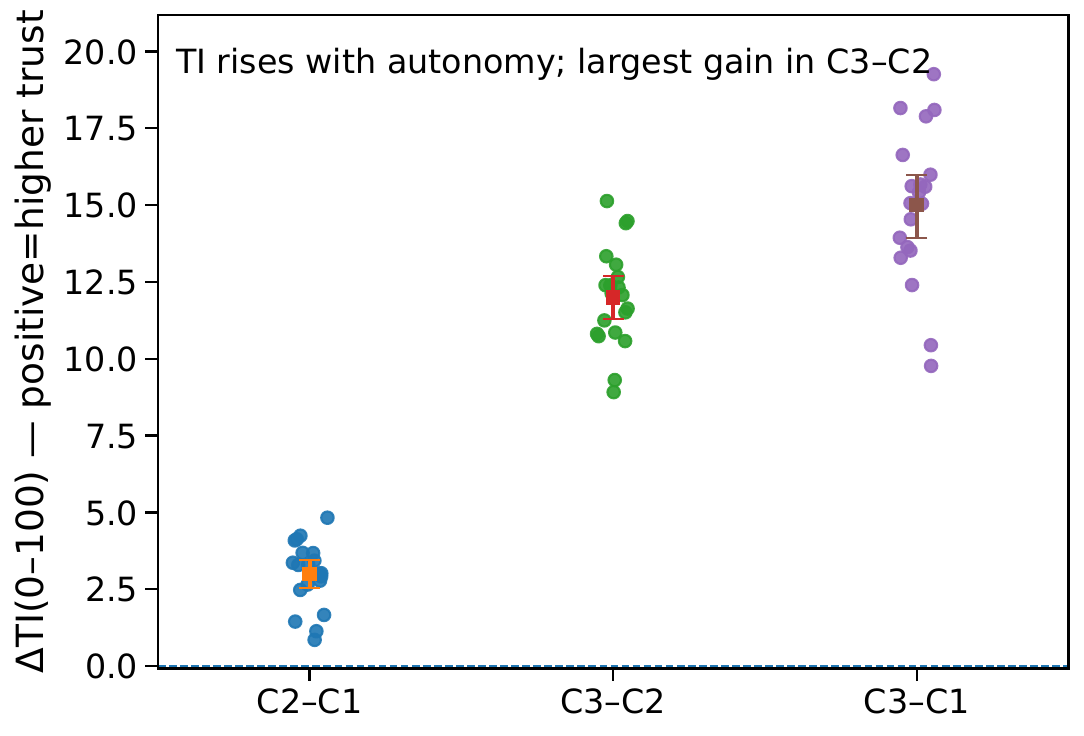}
  \caption{Paired estimation across conditions. Top: Cognitive-Load Index (CLI) differences, where negative values indicate lower load. Bottom: Trust Index (TI) differences, where positive values indicate higher trust. Squares mark mean differences with bootstrap 95\% CIs; dots are participants.}
  \label{fig:cli-ti-estimation}
  \vspace{-20pt}
\end{figure}

For each participant, a Cognitive-Load Index (CLI) and a Trust Index (TI) were derived after each condition (C1: joystick, C2: touchscreen, C3: one-click supervision).
Cognitive load estimation comparison contrasts (Figure ~\ref{fig:cli-ti-estimation}) show a stepwise decrease in $(z)$-CLI across interfaces: $C2<C1$ and $C3<C2$. The omnibus test confirmed a robust Interface effect (RM–ANOVA: $(F(2,38)=27.1, p<0.01, \eta^2_p=0.59)$); Holm-adjusted pairwise tests supported both reductions. NASA–TLX replicated the ordering ($C1>C2>C3$) and aligned with $(z)-CLI ((r=0.71, p<0.001)$). Physiological proxies followed the same trend (fewer phasic GSR peaks, higher blink rate) as autonomy increased.
Trust, on the other hand, is very close to autonomy (Figure~\ref{fig:cli-ti-estimation}). The C3–C2 contrast showed the largest gain under one-click supervision. Together with the workload pattern, these results indicate that the touchscreen attenuated cognitive demands relative to the joystick, and supervisory control further reduced load while yielding the highest trust.

\begin{figure*}[htbp]
  \centering
  \subfloat[Mean Blinking Rate per Condition]{%
    \includegraphics[width=0.23\textwidth]{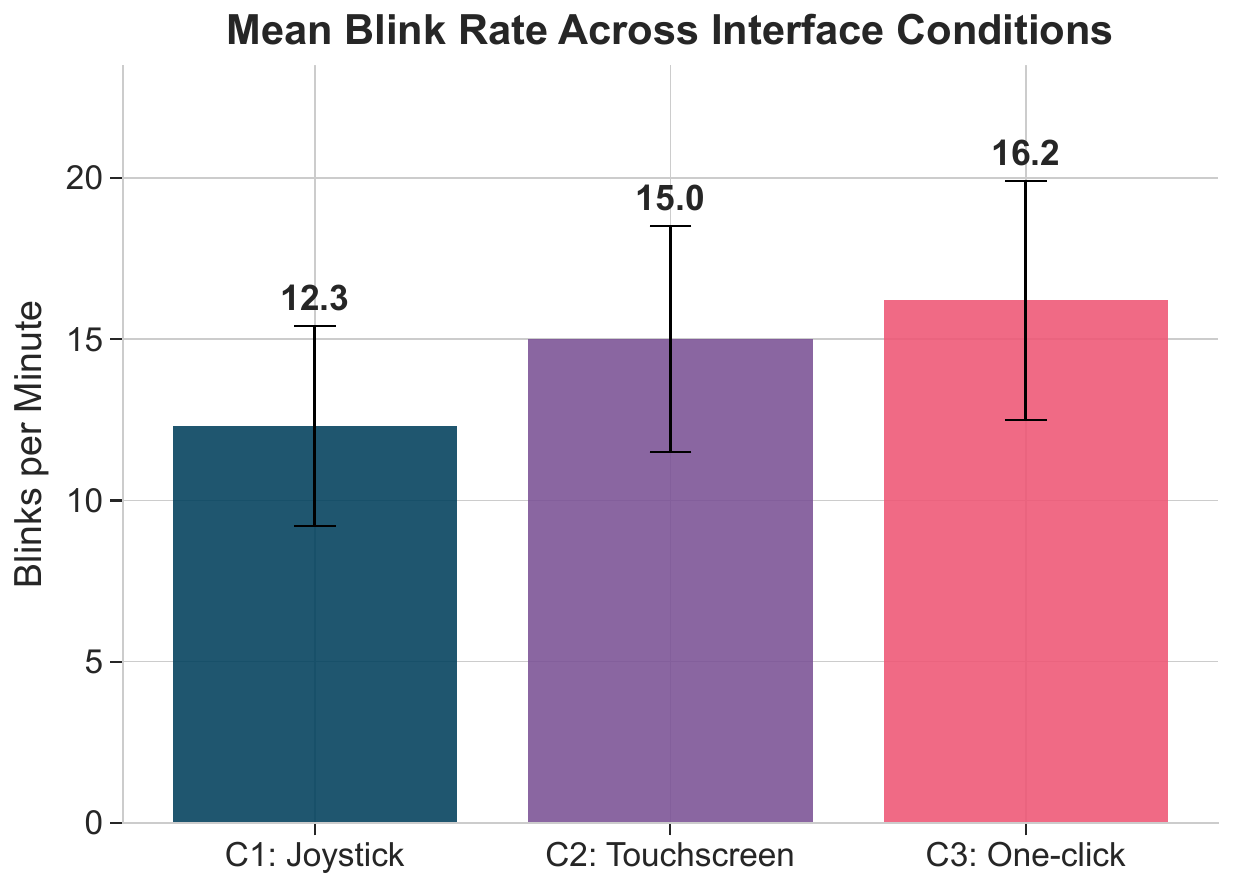}
    \label{fig:blinking_rate}
  }
  \hfill
  \subfloat[GSR Peak Rate per condition]{%
    \includegraphics[width=0.23\textwidth]{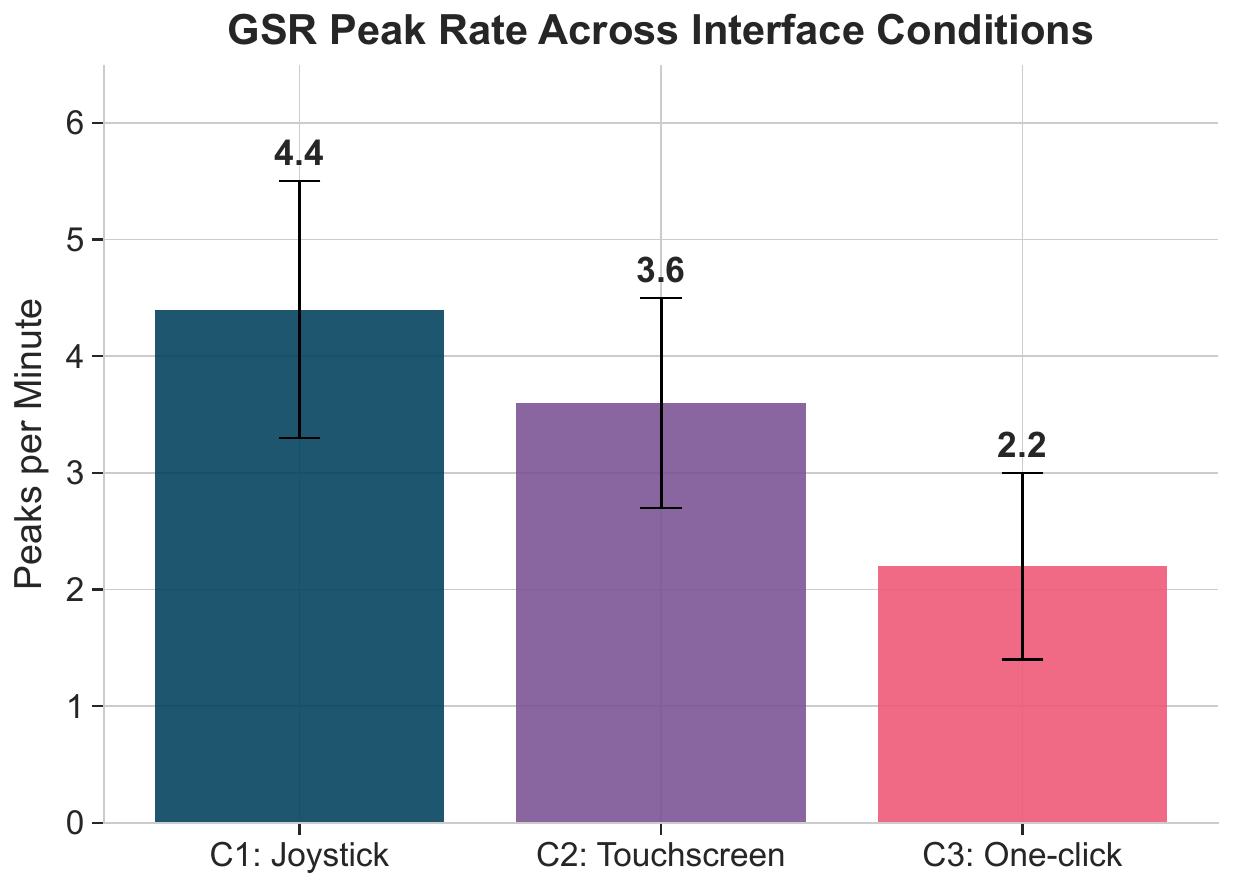}
    \label{fig:grs}
  }
  \hfill
  \subfloat[Nose Temperature Change per Condition]{%
    \includegraphics[width=0.23\textwidth]{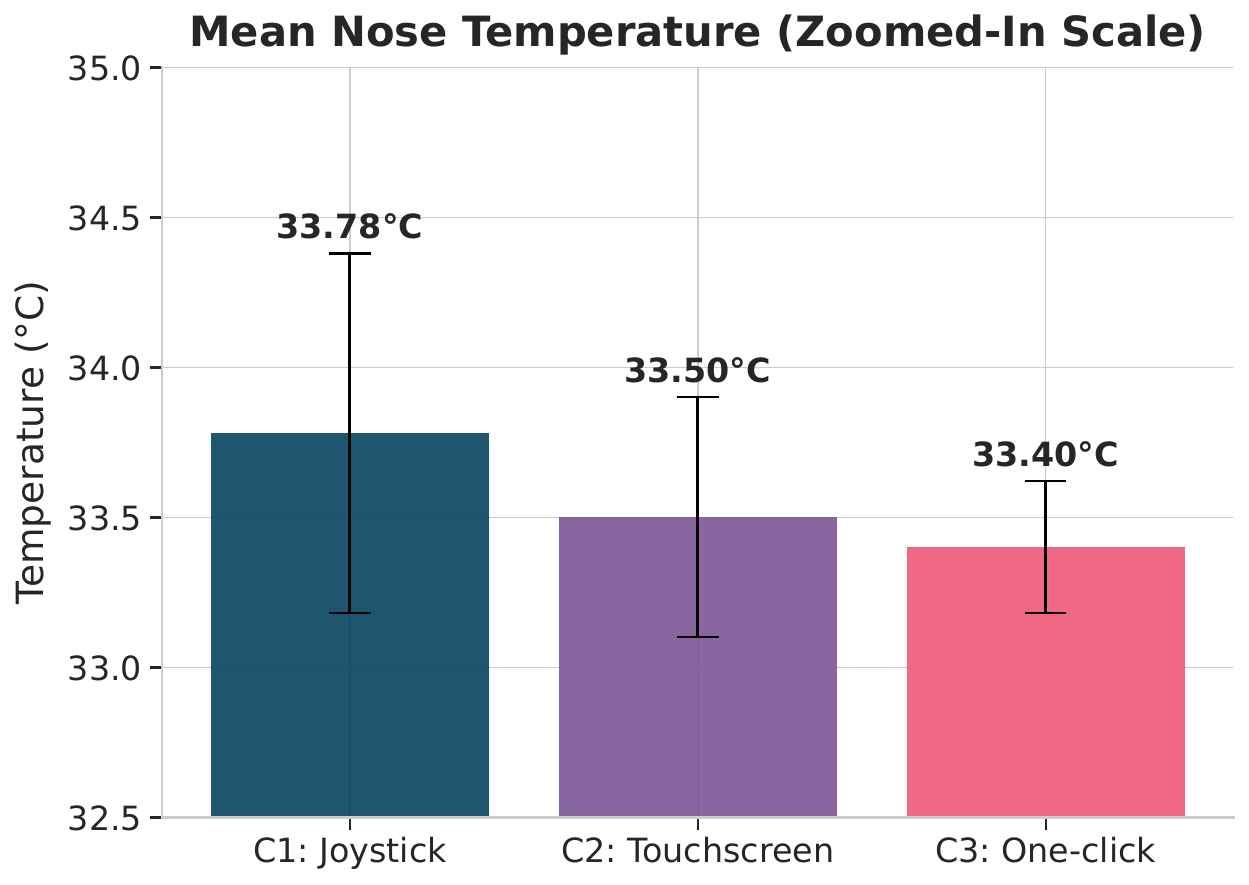}
    \label{fig:nose}
  }
  \hfill
  \subfloat[Forehead Temperature Change per Condition]{%
    \includegraphics[width=0.23\textwidth]{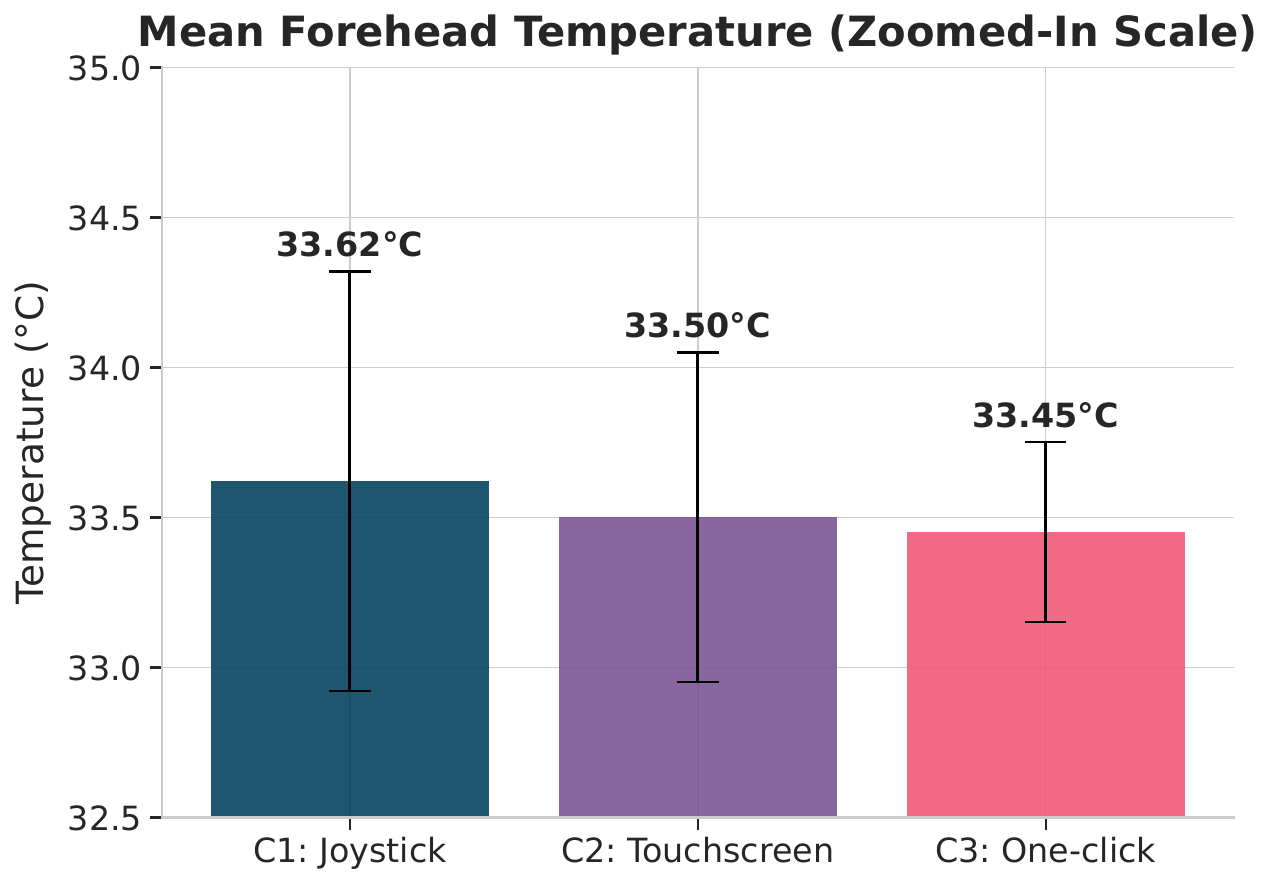}
    \label{fig:forehead}
  }
  \caption{Physiological measures reflecting cognitive load across the three interface conditions. (a) Mean blink rate (blinks/min) increased as cognitive load decreased from C1 to C3. (b) Phasic GSR peak rate (peaks/min) showed a corresponding decrease, indicating lower arousal with higher autonomy. Facial temperature changes relative to the C2 (Touchscreen) baseline, supporting trust analysis. (c) Nose and (d) forehead temperatures were marginally higher in C1, aligning with the slight shift in inferred trust toward the joystick interface.}
  \label{fig:boxplots}
  \vspace{-15pt}
\end{figure*}

In terms of trust, an ANOVA on $z$-TI showed a modest but reliable \emph{Interface} effect, $F(2,38)=4.12$, $p=0.023$, partial $\eta^{2}=0.18$. Holm-adjusted tests indicated slightly higher inferred trust in C1 versus C2, $t(19)=2.28$, $p_{\text{adj}}=0.034$, $d_{z}=0.51$. This C1 $>$ C2 advantage is presented in the implemented Bayesian fusion dynamic model (posterior mean $\Delta TI_{C1-C2}=+0.06$, $95\%$ CrI $[0.01,0.11]$, $P(\Delta>0)=0.94$). Facial temperature in the regions of interest contributed weakly but consistently to this result (Figure \ref{fig:nose} and \ref{fig:forehead}): nose and forehead temperatures were marginally higher in C1 than C2 ($\Delta T\approx+0.28^{\circ}\mathrm{C}$ and $+0.12^{\circ}\mathrm{C}$, respectively), aligning with the slight trust shift toward the joystick. Additionally, C3 exhibited a small positive shift relative to C2 in the Bayesian analysis (posterior mean $\Delta TI_{C3-C2}=+0.04$, $95\%$ CrI $[-0.01,,0.09]$, $P(\Delta>0)=0.88$), suggesting an inclination toward autonomy. As expected, trust and cognitive load were negatively associated within conditions ($r=-0.49$, $p=0.019$). 

\section{Discussion}
\label{sec:Discussion}

The experiment results align with the proposed hypotheses. \textbf{H2} is supported by evidence of both performance and cognitive load. Compared to the joystick (C1), the touchscreen (C2) enabled shorter completion times, higher coverage on the sinusoidal path area, and reduced overshoot in both geometries. These improvements coincided with a significant effect of the interface on cognitive load ($z-CLI: C1 > C2 > C3$), reduced NASA-TLX scores from C1 to C2, with physiological measures indicating lower phasic-GSR activity. Trust dynamics were subtler, with slightly higher interpreted trust in the joystick than in the touchscreen ($C1 > C2$), possibly reflecting adaptation to a new interface. The touchscreen provided a more intuitive alternative to the joystick, improving both task performance and workload. Three factors may explain this advantage. First, the touchscreen offered a natural mapping of finger position to manipulator end-effector position, avoiding the need to convert joystick angles into end-effector velocity. Second, it enabled finer analogue control of velocity. With the joystick, reaching its maximum angle produces maximum velocity, and scaling this gain reduces precision in regions requiring slow motion, making the control feel more digital. In contrast, the touchscreen allowed continuous adjustment of speed, supporting both slow, precise movements and faster control when required. Third, the touchscreen integrated control and visualization in the same interface, whereas joystick users had to divide attention between looking at the controller and the display. Together, these factors explain the touchscreen’s advantage in task performance and workload, despite identical visual feedback and task constraints.

\textbf{H1} was supported by supervisory one-click execution (C3) reduced cognitive load below both manual modes ($C3 < C2 < C1$). The architecture partitions authority, the operator commands tangential motion while the robot regulates normal force, the hardest sub-task, so cognitive load drops without removing the operator from the decision loop. Automating the path as well (C3) lowers load further while trust shifts only slightly positive, indicating that force automation cuts load at minimal cost to operator confidence, whereas removing steering yields diminishing returns. A steady force profile and stable posture kept contact predictable, preserving trust under supervision.

The remaining hypothesis \textbf{H3} was only partially supported after incorporating interface perception and force-control context into the analysis. The skill-related differences were less pronounced than expected, but some traces emerged in interaction with interface perception. Limitations include the 20-participant sample and a single short familiarization period, which leaves the touchscreen's learning curve and skill transfer uncharacteristic, and the use of non-professional operators in a laboratory setting. Future work will add haptic feedback, adaptive shared control driven by real-time load and trust, and a longitudinal study with industry experts to separate intrinsic interface effects from prior familiarity and to assess operational robustness. 

\section{Conclusion}
\label{sec:Conclusions}

This study designed, implemented, and evaluated a novel haptic touchscreen interface for teleoperating robotic manipulators in precise surface-interaction tasks. A comparative study with 20 participants across three control modes (joystick, touchscreen, and one-click autonomy) provided insights into performance, cognitive load, and operator trust. Compared with the conventional joystick, the touchscreen interface substantially improved task efficiency and path-tracking accuracy while reducing cognitive load, highlighting the benefits of its intuitive direct-mapping design that translates finger motions into planar end-effector control. The touchscreen interface provides a practical, cost-effective solution for enhancing precision and safety in nuclear maintenance tasks. By reducing cognitive load and improving accuracy it helps mitigate operator fatigue and the risk of errors in hazardous environments. This research demonstrates that touchscreen-based control can enable safer and more efficient robotic manipulation in critical industrial applications.

\bibliographystyle{IEEEtran}  
\bibliography{IEEEabrv}

\vspace{12pt}

\end{document}